\documentclass[10pt,twocolumn,letterpaper]{article}

\usepackage{3dv}
\usepackage{times}
\usepackage{epsfig}
\usepackage{graphicx}
\usepackage{amsmath}
\usepackage{amssymb}
\usepackage{esvect}
\usepackage{microtype}
\usepackage{enumitem}
\usepackage{pdfpages}

\usepackage{algorithm}
\usepackage{algorithmicx}
\usepackage{algpseudocode}

\usepackage{booktabs}
\usepackage{tabularx}
\newcolumntype{Y}{>{\centering\arraybackslash}X}
\usepackage{xspace}

\usepackage[pagebackref=true,breaklinks=true,letterpaper=true,colorlinks,bookmarks=false]{hyperref}

\threedvfinalcopy 

\def\threedvPaperID{67} 

\ifthreedvfinal\fi
\begin{document}

\title{Rotation Invariant Convolutions for 3D Point Clouds Deep Learning}

\author{
Zhiyuan Zhang$^{1}$ \hspace{0.2in}
Binh-Son Hua$^{2}$ \hspace{0.2in}
David W. Rosen$^{1}$ \hspace{0.2in}
Sai-Kit Yeung$^{3}$
\vspace{0.1in}\\
$^{1}$Singapore University of Technology and Design \quad
$^{2}$The University of Tokyo
\\
$^{3}$Hong Kong University of Science and Technology
\\
}

\maketitle

\def\ournet{RINet\xspace}
\def\ourconv{RIConv\xspace}
\def\smallgap{\vspace{0.05in}}
\newcommand{\new}[1]{\textcolor{red}{#1}}
\begin{abstract}
   Recent progresses in 3D deep learning has shown that it is possible to design special convolution operators to consume point cloud data. However, a typical drawback is that rotation invariance is often not guaranteed, resulting in networks that generalizes poorly to arbitrary rotations.
   In this paper, we introduce a novel convolution operator for point clouds that achieves rotation invariance. Our core idea is to use low-level rotation invariant geometric features such as distances and angles to design a convolution operator for point cloud learning. The well-known point ordering problem is also addressed by a binning approach seamlessly built into the convolution. This convolution operator then serves as the basic building block of a neural network that is robust to point clouds under 6-DoF transformations such as translation and rotation.  
   Our experiment shows that our method performs with high accuracy in common scene understanding tasks such as object classification and segmentation. Compared to previous and concurrent works, most importantly, our method is able to generalize and achieve consistent results across different scenarios in which training and testing can contain arbitrary rotations. Our implementation is publicly available at our project page \footnote{\url{https://hkust-vgd.github.io/riconv/}}.
\end{abstract}

\section{Introduction}
\label{indroduction}
Recent 3D deep learning has led to great progress in solving scene understanding problems like object classification, semantic and instance segmentation with high accuracies by training a neural network with 3D data. Researches in this area have been continuing to grow and diverse as 3D data becomes more widely and easily available from consumer devices.

Among various data presentations, 3D point cloud is a strong candidate for scene understanding tasks thanks to its availability, compactness, and robustness compared to volumetric or image representations. Point clouds can be acquired by various methods and hardware including multiple view geometry in dual-lens cameras and structured light or time-of-flight sensing in depth and LiDAR cameras. However, learning with point clouds is deemed challenging because a point cloud does not contain a regular structure such as that in an image or a volume. Performing convolution on point cloud therefore requires some special designs in the convolution operator that takes care of this irregularity. A wide body of works~\cite{qi2017pointnet,qi2017pointnet++,hua2017point,le2018pointgrid,li2018pointcnn,wang2018edgeconv,xu2018spidercnn,shabat20183dmfv} have recently been proposed to solve this problem, demonstrating state-of-the-art performance in scene understanding with point cloud data. 

Nevertheless, there remains a fundamental problem with existing convolution operator with point clouds: most previous works do not allow the input point cloud to be \emph{rotation invariant}. During training, data is simply augmented with some rotations which can cause the network not able to generalize well to unseen rotations. A few convolution operators that allows rotation invariance exist~\cite{esteves2018learning,rao-spherical-cvpr19} but consistent predictions with arbitrarily rotated data are still not achieved.    

In this work, we propose a novel convolution operator for point clouds that can achieve high accuracies in scene understanding tasks while still preserving the rotation invariance property. Particularly, our convolution is based on low-level geometric features that are translation and rotation invariant. Such features are used in tandem with a binning approach that addresses point ordering issue in point cloud convolution, resulting in a single convolution that is robust to both issues.  
In summary, our contributions are:
\begin{itemize}[leftmargin=*]
\item  A robust feature extraction scheme suitable for convolution that supports both rotation and translation invariant features based on low-level geometric cues; \vspace{-0.1in}

\item  A novel convolution operator that is agnostic to both point cloud rotations and point orders. To address the point ordering issue, we devise a simple binning approach that can be seamlessly combined with the feature extraction step; \vspace{-0.1in}

\item A compact convolutional neural network based on the proposed convolution for object classification and object part segmentation. We demonstrate highly consistent and accurate performance under different rotations.
\end{itemize}

\section{Related Works}
\label{related_works}

The availability of 3D object and scene datasets~\cite{wu-3dshapenets-cvpr15, yi2016scalable,armeni-parsing-cvpr16,hua2016scenenn,dai2017scannet} has made scene understanding in 3D feasible. Common tasks such as object classification, semantic segmentation, and retrieval can now achieve highly accurate results. We briefly summarize the development of 3D deep learning below. 

3D deep learning is more diverse compared to image-based deep learning because there are various representations for learning with 3D data. In early stage of 3D deep learning, volume representation~\cite{wu20153d,maturana-voxnet-iros15,qi2016volumetric,li2016fpnn}, or multiple view images~\cite{su2015multi,qi2016volumetric} are often adopted for neural networks since they are straightforward extensions from learning with images. However, such representations do not scale well due to large memory requirements and limited resolution in representing 3D geometry. 

Recently, PointNet~\cite{qi2017pointnet,qi2017pointnet++} sparked the research interest in deep learning with 3D point clouds by showing that it is possible to learn features of a point set with a special network that is robust to input point orders. This opens the capability for object classification and semantic segmentation with point clouds. Several subsequent works are built along this line of research.
Alternatives to make convolution operator compatible to point cloud is to summarize point features into a regular grid and apply a traditional convolution~\cite{hua2017point,le2018pointgrid}, performing convolution on a local space such as tangent planes~\cite{tatarchenko2018}, learning to transform point clouds into a canonical latent space~\cite{li2018pointcnn}. Such techniques perform competitively to PointNet while being able to exploit features from a local region on the point cloud. 

The trend of deep learning with point cloud data has been continuing to grow diversely. 
Recent methods explores convolution kernels that exploit geometric features~\cite{shen2018mining}, add edges on top of points~\cite{wang2018edgeconv}, parameterize convolution using polynomials~\cite{xu2018spidercnn}, and leverage shape context~\cite{xie2018shapecontext}. Some methods are specially design to be lightweight for real-time applications~\cite{shabat20183dmfv}, or to combine with recurrent neural network~\cite{huang2018recurrent} and sequence model~\cite{liu2018point2seq}. 
Some methods exploit hierarchical structures and clustering for scalability~\cite{riegler2017octnet,klokov2017escape,wang2017cnn,wang2018aocnn,landrieu-superpoint-cvpr18}, mapping point cloud to two dimensional space~\cite{yang2018foldingnet, groueix2018atlasnet,li2018sonet}, applying spectral analysis~\cite{yi2017syncspeccnn}, or addressing non-uniform point distribution~\cite{hermosilla2018monte}.

Our method is a part of this trend. We explore how to perform convolution on local point features and at the same time achieve rotation invariance. Compared to deep learning with images, rotation invariance is an important property and a more critical issue for robustness because in 3D, there is no convention about how to align 3D shapes. 
In geometric deep learning~\cite{bronstein-geometric-ieee17}, one can achieve rotation invariance with geodesic convolution on Riemannian manifolds with angular maxpooling~\cite{masci-geodescic-iccv15w}. Such convolution, however, needs shape surfaces to operate. By contrast, our convolution is for point sets, and defined directly in the Euclidean space.

The most relevant work to ours is the concurrent work by Rao et al.~\cite{rao-spherical-cvpr19}. They showed that point clouds can be mapped to an icosahedral lattice on which a rotation invariance convolution can be implemented. The key difference here is that we do not need a spherical domain for rotation invariance. Instead, we define convolution with rotation invariant features, which is much simpler and intuitive. 
In addition, there are a few previous works about learning local descriptors from point clouds for feature matching~\cite{zeng-3dmatch-cvpr17,khoury-cfg-iccv17,deng-ppffoldnet-eccv18}, some of which~\cite{deng-ppffoldnet-eccv18} can be rotation invariant. These works are however orthogonal to ours mainly because they are targeted for point cloud registration. 


\section{Rotation Invariant Convolution}
\label{sec:conv}
\begin{figure}[b]
	\centering
	\includegraphics[width=\linewidth]{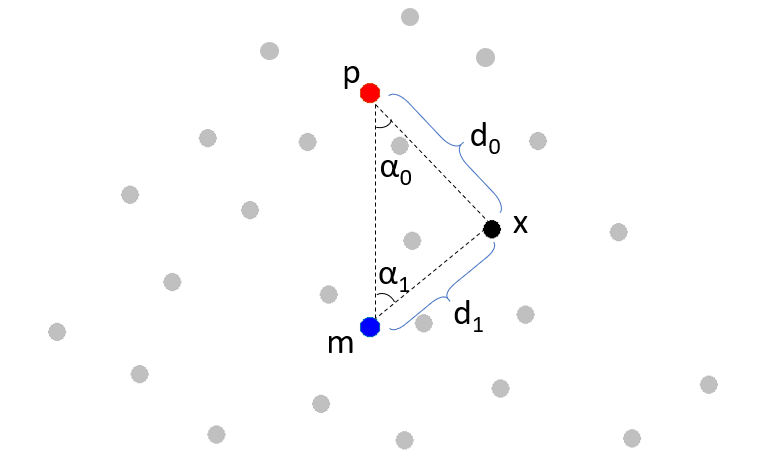}
	\caption{Rotation invariant feature extraction for a point set. At each point (grey), we compute distances and angles to a reference vector established from a reference point (red) and the centroid (blue). Such geometric cues can be directly computed in the Euclidean space, facilitating the design of our convolution operator.}
	\label{fig:rif}
\end{figure}

\begin{figure*}[t]
	\centering
	\includegraphics[width=\linewidth]{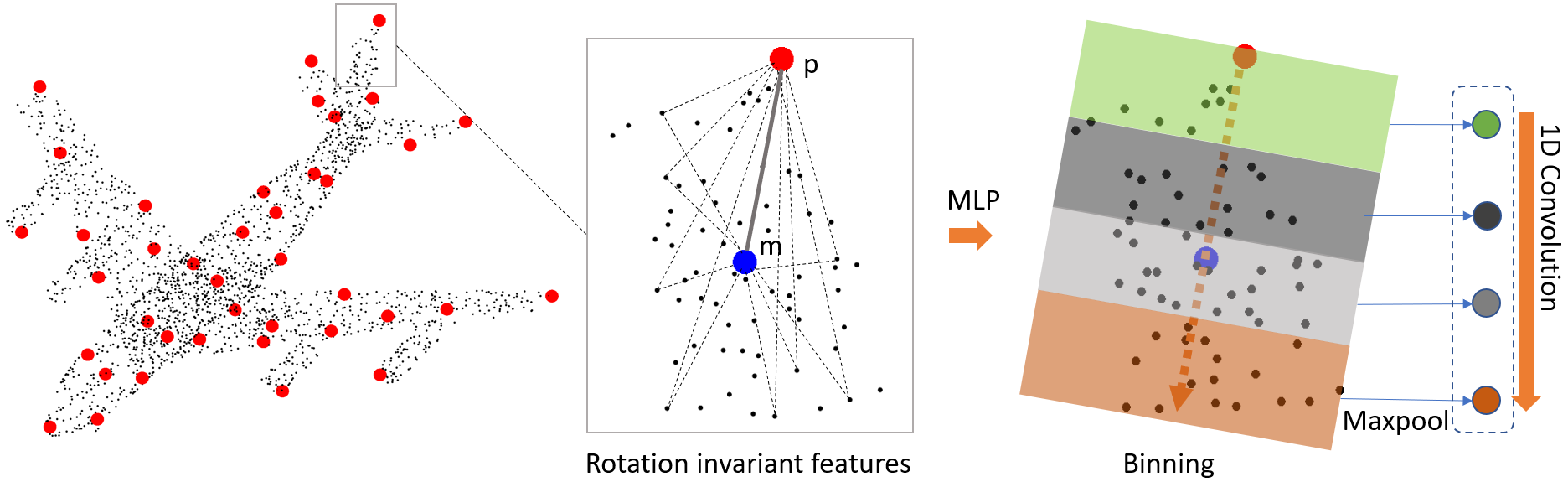}
	\parbox[b]{1\textwidth}{\relax
		\quad \qquad \quad \quad \qquad (a) \quad \qquad \qquad \qquad \quad \quad \qquad \qquad \qquad (b) \qquad \qquad \qquad \qquad \qquad \qquad \qquad(c)}
	\caption{\ourconv operator construction. (a) For an input point cloud with/without associated features, representative points (red dots) are sampled via farthest point sampling. (b) For a reference point $p$, $K$ neighbors are queried to yield a local point set. Also, the centroid of this point set is computed and denoted as $m$ (blue dot). Vector $\protect\vv{pm}$ serves as the reference orientation and the point set is transformed into rotation invariant features using the method described in subsection \ref{sec:RIF}, which is further lifted to a high-dimensional space by a shared multi-layer perceptron (MLP). (c) The local space is then uniformly divided into several bins along $\protect\vv{pm}$ and the points of each bin are summarized by maxpooling. Finally, a 1D convolution can be performed to obtain the final features.}
	\label{fig:rio}
\end{figure*}

In this section, we detail the \ourconv operator construction procedure. Our goal is to seek a simple but efficient way to perform traditional convolution on features extracted from an input point cloud. We design a feature extraction scheme such that the local features are invariant to both translation, rotation, and point orders. Different from previous works that rely on a spherical convolution for rotation invariance~\cite{esteves2018learning}, we show that it is possible to achieve rotation invariance directly in the Euclidean space by utilizing low-level geometric cues.

\subsection{Rotation Invariant Local Features}
\label{sec:RIF}
Our feature extraction can be explained as in Figure~\ref{fig:rif}. Given a reference point $p$ (red), $K$ nearest neighbors are determined to construct a local point set. The centroid of the point set is denoted as $m$ (blue). 
We use vector $\vv{pm}$ as a reference to extract translation and rotation invariant features for all points in the local point set. Particularly, for a point $x$ in this set, its features are defined as
\begin{equation}
	RIF(x ; \vv{pm}) = [d_{0}, d_{1}, \alpha_{0}, \alpha_{1}]\,.
\end{equation}
Here, $d_{0}$ and $d_{1}$ represent the distances from $x$ to $p$ and to $m$, respectively. $\alpha_{0}$ and $\alpha_{1}$ represent the angles from $x$ towards $p$ and $m$, as shown in Figure~\ref{fig:rif}. Since such low-level geometric features are invariant under rigid transformations, they are very well suited for our need to make a translation invariant convolution with rotation invariance property. 
Note that the reference vector $\vv{pm}$ can also serve as a local orientation indicator and we will use it to build a local coordinate system for convolution, in the subsequent step.

A caveat from the feature extraction scheme is that the reference vector $\vv{pm}$ can degenerate when $p$ and $m$ become a single point. Such cases occur when the neighbors are distributed evenly around the reference point. In such case, we select the farthest point to $p$ as $m$ to avoid singularity. In fact, within such a smooth distribution, points that are equidistant to the reference point are expected to have similar features, and thus the degeneration does not negatively affect the features. 

\begin{figure*}[t]
	\centering
	\includegraphics[width=\linewidth]{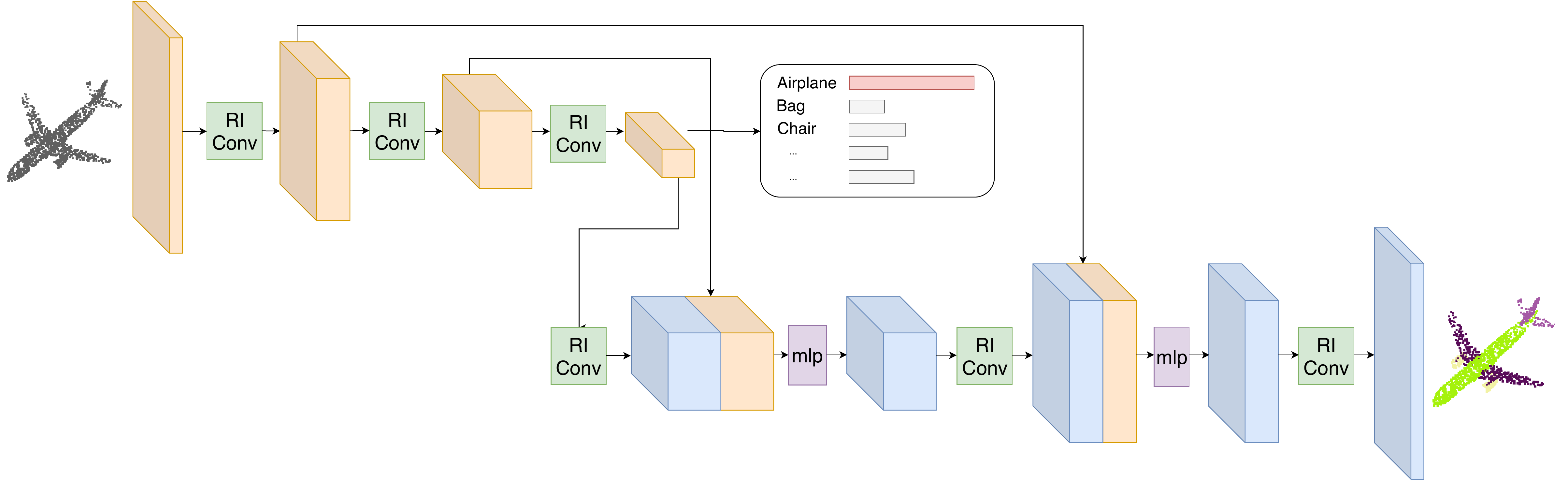}
	\caption{Our proposed network architecture. We follow a convolutional neural network design for classification and part segmentation. Skip connections are further used to combine features from the encoding stage to the decoding stage in part segmentation.}
	\label{fig:rio_net}
\end{figure*}

\subsection{Convolution Operator}
\label{sec:RIConv}
After obtaining rotation invariant features, we are now ready to detail the main idea of our convolution in Figure~\ref{fig:rio}. A key issue here is how to perform convolution that is agnostic to input point orders. PointNet~\cite{qi2017pointnet} extracts a global feature vector from the entire input point cloud by maxpooling the features from a shared MLP. Here, we build our convolution on local features and use a binning approach with shared MLP to solve this issue. This idea is relevant to shell based convolution~\cite{zhang-shellnet-iccv19} in that both apply binning to resolve the ordering issue of point sets and output fixed size features. 

Particularly, we start by sampling a set of representative points through farthest point sampling strategy which is able to generate uniformly distributed points. From each of which we perform a set of K-nearest neighbors to obtain local point sets. For each point, the rotation invariant features are extracted as described in the previous section. The features are lifted to a high-dimensional space by a shared multi-layer perceptron (MLP). 

To proceed with convolution, we have to define an order so that kernel weights in the convolution can be applied to the corresponding points. Here we devise a simple binning approach and turn the convolution into 1D. Such process has been shown to be highly efficient for local feature learning~\cite{zhang-shellnet-iccv19}. In this work, the steps are as follows. We use the reference vector $\vv{pm}$ and split the point distribution into N cells along this vector. The feature of each cell is maxpooled from all points participating in the cell. As the cells are ordered, convolution thus becomes possible. We apply a 1D convolution on the fixed-size feature vector from the cells to obtain the output features of our operator. All steps are summarized in Algorithm~\ref{alg:conv} (see Appendix).

In addition, traditional convolutional neural networks often allows downsampling and upsampling to manipulate the spatial resolution of the input. We build this strategy into our convolution by simply treating the reference point set as the downsampling/upsampling points.

\section{Neural Networks}
\label{sec:net}

We use our convolution operator as the core to build neural networks for two common scene understanding tasks: object classification and object part segmentation. These two tasks are commonly used to benchmark the performance of deep learning with point cloud data~\cite{qi2017pointnet}. Our network is shown in Figure~\ref{fig:rio_net}.

The object classification network consists of three rotation invariant convolution operators followed by a classifier to output labels for the input point cloud. As our convolution operator is already designed to handle arbitrary rotation and point orders, we can simply place each convolution one after another. By default, each convolution is followed by a batch normalization and an ReLU activation. 

The object part segmentation network follows an
encoder-decoder architecture with skip connections similar to U-net~\cite{ron2015unet}. We assume a general condition that the object category is unknown when part segmentation is performed. 
The classification network acts as the encoder, yielding the features in the latent space that can be subsequently decoded into part labels. 

In the decoding stage, after each feature is concatenated by skip connections, we apply a MLP before passing the features for deconvolution. Our deconvolution is basically similar to convolution except that it gradually outputs denser points with less feature channels until the output reaches the original number of points.

\paragraph{Convolution Parameters.} Unless otherwise mentioned, we use 1024 points for classification, and 2048 points for part segmentation, respectively. In the encoding stage, the point cloud is downsampled to $256$, $128$, and $64$, respectively for classification task, and $512$, $128$, and $32$, respectively for segmentation task.
The nearest neighbor size is set to $64$, $32$, and $16$ respectively for the three layers of convolutions. We empirically set the number of bins for handling point orders in each convolution as $4$, $2$, $1$, respectively, which strikes a good balance between accuracy and speed.
This setting ensures that each bin contains $16$ points approximately. In general, the neighborhood has to be large enough for capturing the point distribution and features robustly but not too large that causes too much overhead. 

\section{Experimental Results}
\label{experiment}

\begin{table*}[t]
	\centering
	\footnotesize
	\begin{tabular}{l|lll|ccc|c}
		\toprule
		Method  & Input & Input size & Parameters & z/z & SO3/SO3 & z/SO3 & Acc. std.\\
		\midrule  
		VoxNet \cite{huang2018recurrent} & voxel & $30^{3}$ &0.9M & 83.0 & \textbf{87.3} & - & 3.0 \\ 
		SubVolSup \cite{qi2016volumetric}& voxel & $30^{3}$ &17M  & 88.5 & 82.7 & 36.6 & 28.4 \\ 
		SubVolSup MO \cite{qi2016volumetric}  & voxel & $30^{3}$ &17M & 89.5 & 85.0 & 45.5 & 24.2 \\ 
		Spherical CNN \cite{esteves2018learning}  & voxel & $2\times 64^{2}$ & 0.5M & 88.9 & 86.9 & 78.6 & 5.5\\ 
		MVCNN 12x \cite{su2015multi}  & view & $12\times 224^{2}$ & 99M & 89.5 & 77.6 & 70.1 & 9.8 \\
		MVCNN 80x \cite{su2015multi}  & view & $80\times 224^{2}$ & 99M & 90.2 & 86.0 & 81.5 & 4.3 \\
		PointNet \cite{qi2017pointnet}  & xyz & $1024\times 3$ & 3.5M & 87.0 & 80.3 & 12.8 & 41.0 \\ 
		PointNet++ \cite{qi2017pointnet++}  & xyz & $1024\times 3$ & 1.4M & 89.3 & 85.0 & 28.6 & 33.8\\
		PointCNN \cite{li2018pointcnn}  & xyz & $1024\times 3$ & 0.60M & \textbf{91.3} & 84.5 & 41.2 & 27.2 \\
		\midrule 
		Ours & xyz & 1024 $\times 3$ & 0.70M & 86.5 & 86.4 & \textbf{86.4} & \textbf{0.1}\\
		
		\bottomrule
	\end{tabular}
	\smallgap
	\caption{Comparisons of the classification accuracy (\%) on the ModelNet40 dataset. The accuracy is reported on three test cases: training and testing with z/z, SO3/SO3 and z/SO3 rotation, respectively. Our method has good accuracy and lowest accuracy deviation in all cases.}
	\label{tab:classification}
\end{table*}

We report our evaluation results in this section. We implemented our network in Tensorflow \cite{abadi2016tensorflow}. We use a batch size of $32$ for classification training and $16$ for segmentation training. 
The optimization is done with an Adam optimizer. The initial learning rate is set to 0.001. 
Our training is executed on a computer with an Intel(R) Core(TM) i7-6900K CPU equipped with a NVIDIA GTX 1080 GPU. 

We evaluate the proposed convolution and neural network with two tasks: object classification and object part segmentation. The point cloud size is 1024 for classification and 2048 for segmentation.
It takes about 3 hours for the training to converge for classification, and about 18 hours for part segmentation. Unless otherwise stated, for object classification, we train for $250$ epochs. The network usually converges within $150$ epochs. For object part segmentation, we train for $300$ epochs, and the network usually converges within $200$ epochs.

Following Esteves et al.~\cite{esteves2018learning}, we perform experiments in three cases: (1) training and testing with data augmented with rotation about gravity axis (z/z), (2) training and testing with data augmented with arbitrary SO3 rotations (SO3/SO3), and (3) training with data by z-rotations and testing with data by SO3 rotations (z/SO3). The first case is commonly adopted by previous methods in handling rotated point clouds, and the last two cases are for evaluating rotation invariance. In general, it is expected that a convolution with rotation invariance should generalize well in case (3) even though the network is not trained with data augmented with SO3 rotations.

In general, our result demonstrates the effectiveness of the rotation invariant convolution we proposed. Our networks yield very \emph{consistent} results despite that our networks are trained with a limited set of rotated point clouds and tested with arbitrary rotations. To the best of our knowledge, there is no previous work for point cloud learning that is able to achieve the same level of consistency despite that some methods~\cite{esteves2018learning} demonstrated good performance when trained with a particular set of rotations. We detail our evaluations below. 

\subsection{Object Classification}
\label{sec:eval_classification}
The classification task is trained on the ModelNet40 variant of the ModelNet dataset~\cite{wu20153d}. ModelNet40 contains CAD models from 40 categories such as airplane, car, bottle, dresser, etc. By following Qi et al.~\cite{qi2017pointnet}, we use the preprocessed $9,843$ models for training and $2,468$ models for testing. The input point cloud size is 1024, with each point represented by $(x,y,z)$ coordinates in the Euclidean space.

We followed Li et al.~\cite{li2018pointcnn} and use multiple feature vectors to train the classifier. Particularly, our network outputs $64$ feature vectors of length $512$ to the classifier. Each of these vectors is passed through an $mlp$ implemented by fully connected layers, resulting in $64 \times 40$ category predictions. During training, we apply cross entropy loss to all such predictions. During testing, we take the mean of such predictions to obtain the final category prediction. In Section~\ref{sec:eval_design}, we further evaluate this strategy and show that it leads to better performance than networks with a single feature vector.

The evaluation results are shown in Table~\ref{tab:classification}. Following the work of \cite{esteves2018learning}, we perform experiments in three cases: training and testing with data rotated about the gravity axis (z/z), training and testing with arbitrary SO3 rotations (SO3/SO3), and training with z-rotations and testing with SO3 rotations (z/SO3). The first case is commonly adopted by previous methods in handling rotated point clouds, and the last two cases are for evaluating rotation invariance. 

We use two criteria for evaluation: accuracy and accuracy standard deviation. Accuracy is a common metric to measure the performance of the classification task. In addition, accuracy deviation measures the consistency of the accuracy scores in three tested cases. In general, it is expected that methods that are rotation invariant should be insusceptible to the rotation used in the training and testing data and therefore has a low deviation in accuracy.
 
As can be seen, our method performs favorably to the state-of-the-art techniques. On one hand, our method achieves very good accuracy in all cases despite that there are no clear winner for all cases in our experiment. On the other hand, and more importantly, our method has the lowest accuracy deviation. Previous methods exhibit large accuracy deviations especially in the extreme z/SO3 case. This case is exceptionally hard for methods that rely on data augmentation to handle rotations~\cite{qi2017pointnet,qi2017pointnet++}. In our observation, such techniques are only able to generalize within the type of rotation they are trained with, and generally fail in the z/SO3 test. This applies to both voxel-based and point-based learning techniques. By contrast, our method has almost no performance difference in three test cases, which confirms the robustness of the rotation invariant geometric cues in our convolution. 
We also evaluate the accuracy of the classification task per object category. 
Please see the full results in the supplemental document. 

\paragraph{Network Parameters.} The capability to handle rotation invariance also has a great effect on the number of network parameters. For networks that rely on data augmentation to handle rotations, it requires more parameters to `memorize' the rotations. Networks that are designed to be rotation invariant, such as spherical CNN~\cite{esteves2018learning} and ours, have very compact representations. 
In terms of number of trainable parameters, our network has 0.70 millions (0.70M) of trainable parameters, which is the most compact network in our evaluations. Among the tested methods, only spherical CNN~\cite{esteves2018learning} (0.5M) and PointCNN (0.6M) have similar compactness. Our network has $5\times$ less parameters than PointNet (3.5M), about $2\times$ less than PointNet++ (1.4M). The well balance between trainable parameters, accuracy and accuracy deviations makes our method more robust for practical use.


\subsection{Object Part Segmentation}
\begin{table}[t!]
\footnotesize
	\begin{center}
		\begin{tabular}{l|l|cc}
			\toprule
			Method   & input & SO3/SO3 & z/SO3  \\
			\midrule
			PointNet \cite{qi2017pointnet}     & xyz        & 74.4   & 37.8  \\
			PointNet++ \cite{qi2017pointnet++} & xyz+normal & \textbf{76.7}   & 48.2  \\
			PointCNN \cite{li2018pointcnn}     & xyz        & 71.4   & 34.7  \\
			DGCNN \cite{wang2018edgeconv}      & xyz        & 73.3   & 37.4  \\
			SpiderCNN \cite{xu2018spidercnn}   & xyz+normal & 72.3   & 42.9  \\
			\midrule
			Ours                               & xyz &  75.5 &  \textbf{75.3} \\
			\bottomrule
		\end{tabular}
		\smallgap
		\caption{Comparisons of object part segmentation performed on ShapeNet dataset~\cite{chang2015shapenet}. The mean per-class IoU (mIoU, \%) is used to measure the accuracy under two challenging rotation modes: SO3/SO3 and z/SO3.}
		\label{tab:segmentation}
	\end{center}
\end{table}
We also evaluated our method with the object part segmentation task that aims to predict the part label for each input point.
In this task, we train and test with the ShapeNet dataset~\cite{chang2015shapenet} that contains $16,880$ CAD models in $16$ categories. Each model is annotated with $2$ to $6$ parts, resulting in a total of $50$ object parts. We follow the standard train/test split with $14,006$ models for training and $2,874$ models for testing, respectively. 

\begin{figure*}[t]
	\centering
	\includegraphics[width=\linewidth]{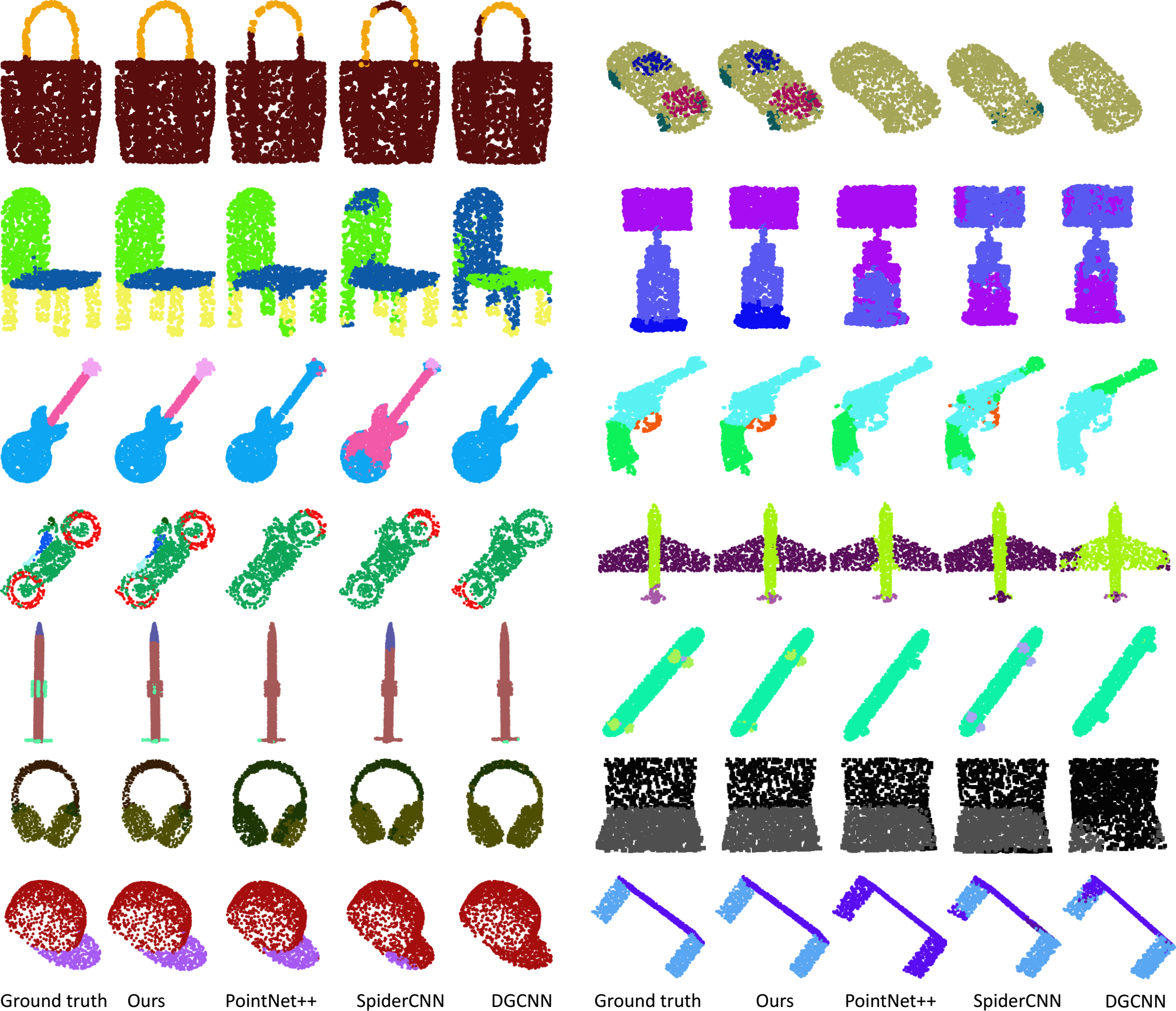}
	\smallgap
	\caption{Qualitative results of object part segmentation task with z/SO3 scenario. Our method has the state-of-the-art performance while previous methods fail to generalize to SO3 rotations.}
	\label{fig:objectpart}
\end{figure*}

The evaluation results are shown in Table~\ref{tab:segmentation}.  
As can be seen, our method outperforms previous methods significantly in z/SO3 test case and achieves similar performance in SO3/SO3 case. This result aligns well with the performance reported in the object classification task. Our method also has consistent performance for both rotation cases, which empirically confirms the rotation invariance in our convolution.
Visualization of our prediction and the ground truth object parts are shown in Figure~\ref{fig:objectpart}. It is easy to observe that our predictions are the closest to the ground truth.
Table~\ref{tab:objectpart_perclass_so3_so3} and Table~\ref{tab:objectpart_perclass_z_so3} further report per-class accuracies for both SO3/SO3 and z/SO3 case. Our method performs best in 3 out of 16 categories in SO3/SO3 case, and 15 out of 16 categories in z/SO3 case. 

\subsection{Evaluations of Network Designs}
\label{sec:eval_design}
In this section, we perform experiments on object classification to analyze the performance and justify the design of the proposed convolution operator and network architecture. Inspired by the fact that there are negligible improvement after $150$ epochs of training (Section~\ref{sec:eval_classification}), we only train the networks with $160$ epochs in this ablation study.

\paragraph{Ablation Study.}
We first experiment by turning on/off different components in our network. The result of this experiment is shown in Table~\ref{tab:ablation}. 
In this table, the Base column indicates a simple network similar to that in Figure~\ref{fig:rio_net} but only contains \ourconv operators to extract local features for classification. 
The MLP indicates the use of an MLP layer to lift rotation invariant features to a high-dimensional feature space. The next two columns indicate the geometric attributes used in \ourconv. The last row shows that when all components are used, we achieve the best accuracy of $86.5\%$. Without high-dimensional feature learning by MLP (first row), the performance drops by almost $3\%$. If we either use angle or distance features (second and third row), the accuracy also drops about $1\%$. This confirms that our network architecture is plausible and yield good performance.

\newcommand{\tabincell}[2]{\begin{tabular}{@{}#1@{}}#2\vspace{-0.15in}\end{tabular}}
\begin{table}[t]
	\footnotesize
	\centering
	\begin{tabularx}{\linewidth}{YYYYY}
		\toprule
		Base  & MLP & Distance & Angle & Acc. \\
		\midrule
		
		\tabincell{c}{\checkmark\\\checkmark \\ \checkmark \\ \checkmark}

		& \tabincell{c}{ \\ \checkmark \\ \checkmark \\ \checkmark} 
		
		& \tabincell{c}{\checkmark \\ \\  \checkmark \\ \checkmark}
		
		& \tabincell{c}{\checkmark \\ \checkmark \\ \\  \checkmark}

		& \tabincell{c}{83.4 \\ 84.6 \\ 84.8 \\ 86.5}\\
		
		\bottomrule
	\end{tabularx}
	\smallgap
	\caption{Ablation study of our method. The results show that combining low-level geometric features such as distances and angles lead to better performance. Besides, using MLP for higher dimensional feature learning can also considerably boost the performance.}
	\label{tab:ablation}
\end{table} 
\newcolumntype{s}{>{\hsize=.33\hsize}X}
\begin{table}[t]
	\footnotesize
	\begin{tabularx}{\linewidth}{s llll}
		\toprule
		\# Layers  & 1  & 2 & 3 & 4 \\
		\midrule
		Accuracy (\%) & 46.8 & 78.2 & 86.5 & 86.8 \\
		Time per epoch (s) & 43.8 & 59.5 & 74.5 & 138.7 \\
	\end{tabularx}
	\begin{tabularx}{\linewidth}{s llll}
		\\
	\end{tabularx}
	\begin{tabularx}{\linewidth}{s llll}
		\toprule
		\# Points  & 128  & 256 & 512 & 1024 \\
		\midrule
		Accuracy (\%) & 76.0 & 80.8 & 84.4 & 86.5 \\
	\end{tabularx}
	\begin{tabularx}{\linewidth}{s llll}
		\\
	\end{tabularx}
	\begin{tabularx}{\linewidth}{s ll}
		\toprule
		\# Features  & Multiple     & PointNet style~\cite{qi2017pointnet} \\
		\midrule
		Accuracy (\%)     & 86.5 & 84.8 \\
		\bottomrule
	\end{tabularx}
	\smallgap
	\caption{Classification accuracy (\%) with different number of convolution layers, input points, and features for classifiers.}
	\label{tab:ablation_layers}
\end{table}

\begin{table*}[t]
	\centering
	\footnotesize
	\begin{tabular}{l p{14pt}p{14pt}p{14pt}p{14pt}p{14pt}p{14pt}p{14pt}p{14pt}p{14pt}p{14pt}p{14pt}p{14pt}p{14pt}p{14pt}p{14pt}p{14pt}}
		\toprule
		Network   & aero & bag & cap & car & chair & earph. & guitar & knife & lamp & laptop & motor & mug & pistol & rocket & skate & table \\ 
		\midrule 
		PointNet \cite{qi2017pointnet}  & \textbf{81.6} & 68.7 & 74.0 & 70.3 & 87.6 & 68.5 & \textbf{88.9} & 80.0  & 74.9 & 83.6 & 56.5 & 77.6 & 75.2 & \textbf{53.9} & \textbf{69.4} & 79.9   \\ 
		PointNet++ \cite{qi2017pointnet++}  & 79.5 & 71.6 & \textbf{87.7} & \textbf{70.7} & \textbf{88.8} & 64.9 & 88.8 & 78.1 & 79.2 & \textbf{94.9} & 54.3 & \textbf{92.0} & 76.4 & 50.3 & 68.4 & 81.0   \\
		PointCNN \cite{li2018pointcnn}  & 78.0 &80.1 &78.2 &68.2 & 81.2 & 70.2 &82.0 &70.6 &68.9 & 80.8 &48.6 &77.3 &63.2 &50.6 &63.2 & \textbf{82.0} \\
		DGCNN \cite{wang2018edgeconv}  & 77.7 & 71.8 & 77.7 & 55.2 & 87.3 & 68.7 & 88.7 & \textbf{85.5} &\textbf{81.8} & 81.3 & 36.2 & 86.0 & 77.3 & 51.6 & 65.3 & 80.2 \\  
		SpiderCNN \cite{xu2018spidercnn}  & 74.3 & 72.4 & 72.6 & 58.4 & 82.0 & 68.5 & 87.8 & 81.3 & 71.3 & 94.5 & 45.7 & 88.1 & \textbf{83.4} & 50.5 & 60.8 & 78.3  \\
		Ours  & 80.6 & \textbf{80.2} & 70.7 & 68.8 & 86.8 & \textbf{70.4} & 87.2 & 84.3 & 78.0 & 80.1 & \textbf{57.3} & 91.2 & 71.3 & 52.1 & 66.6 & 78.5 \\
		\bottomrule
	\end{tabular}
	\smallgap
	\caption{Per-class accuracy of object part segmentation on the ShapeNet dataset in SO3/SO3 scenario. Our method works equally well to previous methods in this scenario.}
	\label{tab:objectpart_perclass_so3_so3}
\end{table*}

\begin{table*}[t]
	\centering
	\footnotesize
	\begin{tabular}{l p{14pt}p{14pt}p{14pt}p{14pt}p{14pt}p{14pt}p{14pt}p{14pt}p{14pt}p{14pt}p{14pt}p{14pt}p{14pt}p{14pt}p{14pt}p{14pt}}
		\toprule
		Network   & aero & bag & cap & car & chair & earph. & guitar & knife & lamp & laptop & motor & mug & pistol & rocket & skate & table \\ 
		\midrule 
		PointNet \cite{qi2017pointnet}  & 40.4 & 48.1 & 46.3 & 24.5 & 45.1 & 39.4 & 29.2 & 42.6  & 52.7 & 36.7 & 21.2 & 55.0 & 29.7 & 26.6 & 32.1 & 35.8   \\ 
		PointNet++ \cite{qi2017pointnet++}  & 51.3 & 66.0 & 50.8 & 25.2 & 66.7 & 27.7 & 29.7 & 65.6 & 59.7 & 70.1 & 17.2 & 67.3 & 49.9 & 23.4 & 43.8 & 57.6   \\
		PointCNN \cite{li2018pointcnn}  &21.8 & 52.0 &52.1 &23.6 &29.4  &18.2 &40.7 &36.9 &51.1 &33.1 &18.9 &48.0 &23.0 &27.7 &38.6 &39.9 \\
		DGCNN \cite{wang2018edgeconv} & 37.0 & 50.2 & 38.5 & 24.1 & 43.9 & 32.3 & 23.7 &48.6 &54.8 & 28.7 & 17.8 & 74.4 & 25.2 & 24.1 & 43.1 & 32.3 \\ 
		SpiderCNN \cite{xu2018spidercnn}  & 48.8 & 47.9 & 41.0 & 25.1 & 59.8 & 23.0 & 28.5 & 49.5 & 45.0 & \textbf{83.6} & 20.9 & 55.1 & 41.7 & 36.5 & 39.2 & 41.2  \\
		Ours  & \textbf{80.6} & \textbf{80.0} & \textbf{70.8} & \textbf{68.8} & \textbf{86.8} & \textbf{70.3} & \textbf{87.3} & \textbf{84.7} & \textbf{77.8} & 80.6 & \textbf{57.4} & \textbf{91.2} & \textbf{71.5} & \textbf{52.3} & \textbf{66.5} & \textbf{78.4} \\
		\bottomrule
	\end{tabular}
	\smallgap
	\caption{Per-class accuracy of object part segmentation on the ShapeNet dataset in z/SO3 scenario. Our method significantly outperforms previous methods thanks to the rotation invariance features from our convolution operators.}
	\label{tab:objectpart_perclass_z_so3}
\end{table*}

\paragraph{Number of Layers.}
We vary the number of convolution layers as follows.
Let us denote the convolution layers in our network in Figure~\ref{fig:rio_net} with $L_{0}$, $L_{1}$, $L_{2}$ from left to right. Here we compare our current architecture with those that have $L_2$ or $L_1$ and $L_2$ removed, or have an additional convolution $L_{-1}$ added before $L_0$. Note that we skip point sampling in $L_{-1}$ to keep the same number of input points. The results in Table~\ref{tab:ablation_layers} (first section) shows the accuracy when the number of layers vary from 1 to 4. 
We can see that with only 1 layer, the accuracy drops dramatically to 46.8\%, which means a single convolution cannot extract effective features. With more convolutions, the accuracy is improved but this comes with the cost of longer training time. Thus, in this work, we choose the architecture of 3 layers for best speed and accuracy balance.

\paragraph{Number of Input Points.} 
We evaluated our network with point clouds of input sizes from $128$ to $1024$ points. Particularly, we retrained and tested the network with point clouds of corresponding number of points. The results are shown in Table~\ref{tab:ablation_layers} (middle section). 
It shows that our network generalizes well to different input size. 

\paragraph{Number of Features for Classifiers.}
For object classification, our network outputs $64$ vectors of length $512$ to the classifier. We compared this strategy with the one in PointNet~\cite{qi2017pointnet} which only outputs a single vector of 512 by maxpooling all features of all points. The results in Table~\ref{tab:ablation_layers} (last section) shows that more output feature vectors yield slightly higher accuracy. Such boost is due to the fact that multiple vectors can convey richer features from different latent spaces that facilitate feature clustering in the classifier.

%

\subsection{Limitations}
Our method is not without limitations. First, the geometric features we used is by no means complete. It is possible to use other more sophisticated low-level geometry features such as curvature to design the convolution. Second, while our convolution is robust and consistent to arbitrary rotations, when there is no rotation or simple rotations as in the z/z case in the classification task, our method is less accurate compared to state-of-the-art classification. This is because the original point coordinates are not retained in low-level geometric feature extraction, trading some discriminative features for rotation invariance. 

We perform an additional experiment in which we remove the proposed geometric features, and replace them with the original 3D coordinates of the input point cloud. This makes our convolution no longer robust to SO3 rotations but in return, the convolution features are more discriminative. This allows us to achieve state-of-the-art accuracy (91.8\% overall accuracy) in the classification task. Fusing original coordinates and geometric features into the same feature space would be therefore a very interesting extension to this work. 

\begin{algorithm*}[t]
	\small
	\caption{\ourconv operator.}
	\label{alg:conv}
	\begin{algorithmic}[1]
		\Require
		Reference point $p$, point set $P$, current point features $F_{prev}$;
		\Ensure
		Convoluted features $F$;
		\State $m \leftarrow avg(P)$;  \hfill * Compute the centroid of $P$
		\State $\vv{pm} \leftarrow m-p$;  \hfill * Determine the reference orientation (Section~\ref{sec:RIConv})
		\State $f_{r} \leftarrow \{ RIF(x ; \vv{pm} ) : \forall x \in P \}$; \hfill * Find rotation invariant features (Section~\ref{sec:RIF})
		\State $F_{r} \leftarrow mlp(f_{r})$; \hfill * Transform each feature $f_{r}$ to high-dimensional feature $F_r$
		\State $F_{in} \leftarrow [F_{prev}, F_{r}]$; \hfill * Concatenate the local and previous layer features
		\State $\{S\} \leftarrow P$; \hfill * Divide local space into $s$ bins along $\vv{pm}$
		\State $\{F_{pool}\} \leftarrow \{ \mathrm{maxpool}(\{ F_{in}(x) : \forall x \in s \}) : \forall s \in S \}$ \hfill * Compute max pooling features for each bin of $\{S\}$
		\State $F \leftarrow \textrm{conv}(\{F_{pool}\})$; \hfill * 1D convolution of the bin features
		\\
		\Return $F$;
	\end{algorithmic}
\end{algorithm*}

\section{Conclusion}
\label{conclusion}
We presented a novel convolution operator for point cloud feature learning that can handle point clouds with arbitrary rotations. Given a point set as input, we determine a reference orientation based on a reference point and the centroid, from which rotation invariant features built upon geometric cues such as distances and angles can be constructed for each point. Combined with a binning strategy, our method handles both rotation invariance and point order issue in a single convolution. 
We then built a simple yet effective end-to-end convolutional neural network for point cloud classification and segmentation. Experiments demonstrate that our method achieves good performance on both classification and segmentation tasks with the best consistency with arbitrary rotation test cases. This is in contrast to existing methods that often perform quite inconsistently for different types of rotations.

Our method leads to several potential future researches. First, the low-level rotation invariance features for convolution are hand-crafted, which we aim to generalize by applying unsupervised learning to learn such features. Second, our convolution could be beneficial to more scene understanding applications such as object detection and retrieval. It would be also of great interest to extend our method to achieve invariance to rigid and non-rigid transformations. 

\noindent
\textbf{Acknowledgement.} 
The authors acknowledge support from the SUTD Digital Manufacturing and Design Centre (DManD) funded by the Singapore National Research Foundation. This project is also partially supported by Singapore MOE Academic Research Fund MOE2016-T2-2-154 and Singapore NRF under its Virtual Singapore Award No. NRF2015VSGAA3DCM001-014.


{\small
\bibliographystyle{ieee}
\bibliography{egbib}
}

\clearpage
\thispagestyle{empty}
\includepdf[pages={1}]{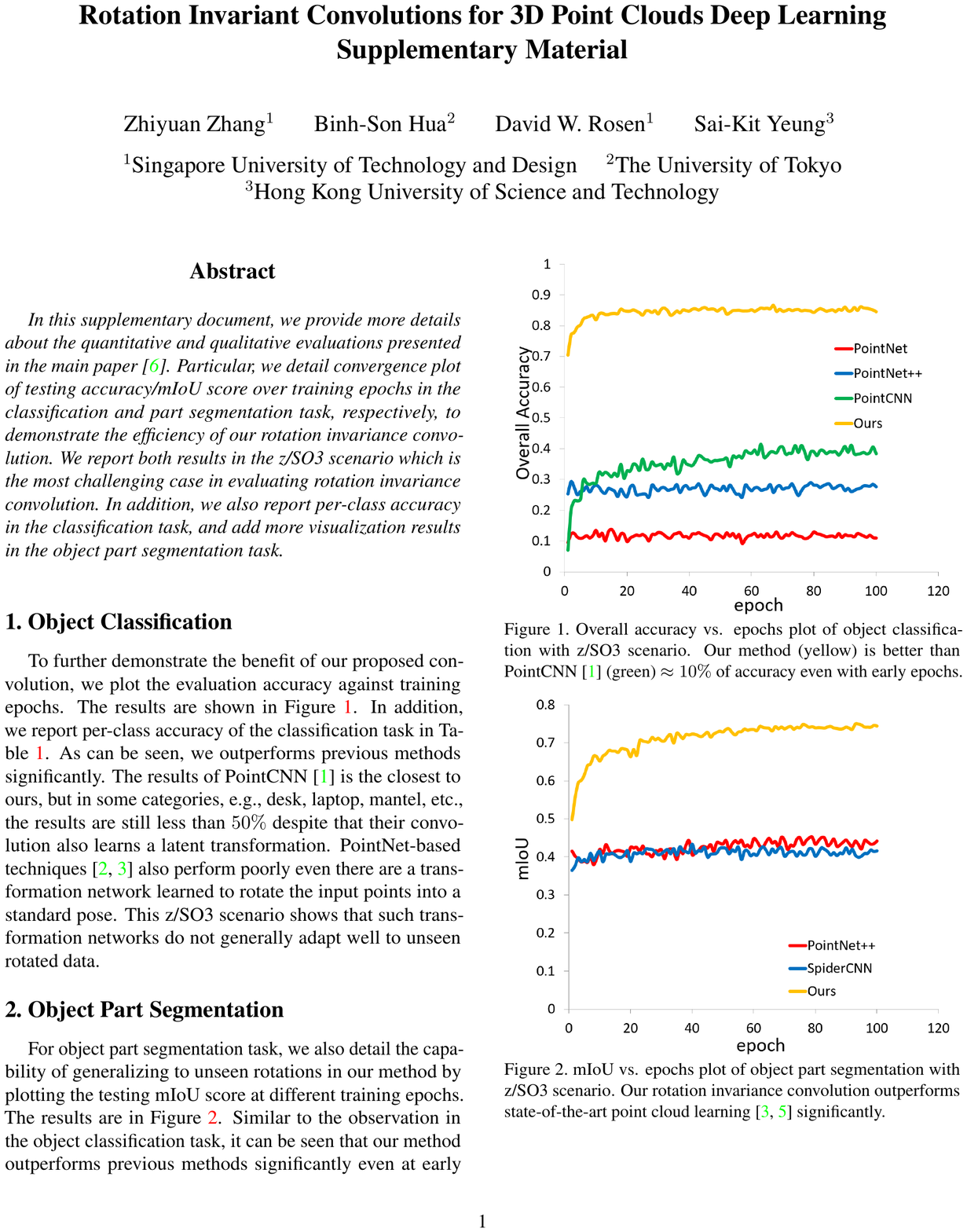}
\includepdf[pages={2}]{supp.pdf}
\includepdf[pages={3}]{supp.pdf}

\end{document}